\documentclass{article}
    \PassOptionsToPackage{numbers, compress}{natbib}

    \usepackage[preprint]{neurips_2019}

\usepackage[utf8]{inputenc} 
\usepackage[T1]{fontenc}    
\usepackage{subfigure}
\usepackage{hyperref}       
\usepackage{url}            
\usepackage{booktabs}       
\usepackage{graphicx}
\usepackage{enumitem}
\usepackage{amsmath}
\usepackage{amsfonts}       
\usepackage{nicefrac}       
\usepackage{microtype}      
\usepackage{algorithm}
\usepackage{algorithmicx}
\usepackage{algpseudocode}
\usepackage{bbding}
\usepackage{graphicx}
\newcommand{\eat}[1]{}
\title{Segmentation Is All You Need}
\author{Zehua Cheng$^{1,2,*}$, Yuxiang Wu$^{3,}$\thanks{Co-first authors, contributed to this work equally.}, Zhenghua Xu$^{1,2}$\thanks{Corresponding author, email:zhenghua.xu@hebut.edu.cn.}, Thomas Lukasiewicz$^2$, Weiyang Wang$^4$\\
$^1$State Key Laboratory of Reliability and Intelligence of Electrical Equipment,\\
Hebei University of Technology, China\\
$^2$Department of Computer Science, University of Oxford, United Kingdom\\
$^3$AISA Research, Hunan Agricultural University, China\\
$^4$SnowCloud.ai, China\\}

\begin{document}
\maketitle
\begin{abstract}
Region proposal mechanisms are essential for existing deep learning approaches to object detection in images. Although they can generally achieve a good detection performance under normal circumstances, their recall in a scene with extreme cases is unacceptably low. This is mainly because bounding box annotations contain much environment noise information, and non-maximum suppression (NMS) is required to select target boxes. Therefore, in this paper, we propose the first anchor-free and NMS-free object detection model, called weakly supervised multimodal annotation segmentation (WSMA-Seg), which utilizes segmentation models to achieve an accurate and robust object detection without NMS. In WSMA-Seg, multimodal annotations are proposed to achieve an instance-aware segmentation using weakly supervised bounding boxes; we also develop a run-data-based following algorithm to trace contours of objects. In addition, we propose a multi-scale pooling segmentation (MSP-Seg) as the underlying segmentation model of WSMA-Seg to achieve a more accurate segmentation and to enhance the detection accuracy of WSMA-Seg. Experimental results on multiple datasets show that the proposed WSMA-Seg approach outperforms the state-of-the-art detectors.

\end{abstract}

\section{Introduction}

\begin{figure}[!ht]
  \centering
  \subfigure[Complex occlusion \label{subfig:occlusion}]{
     \includegraphics[width=0.25\textwidth]{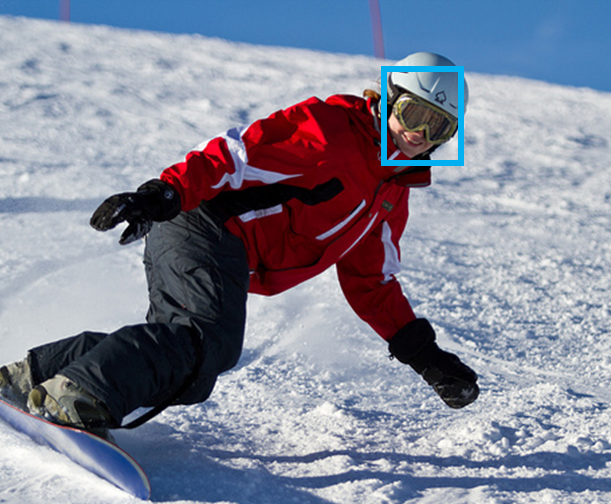}
  }
  \subfigure[Poor illumination \label{subfig:lightning}
  ]{
     \includegraphics[width=0.19\textwidth]{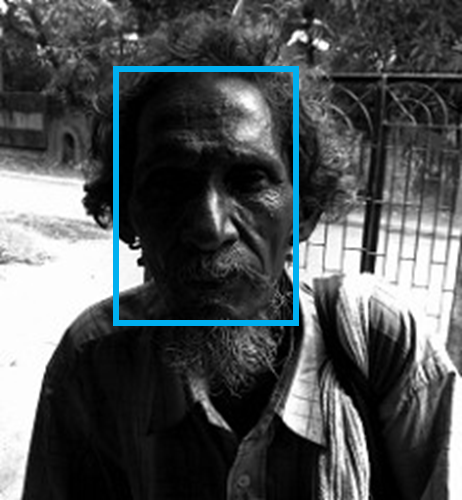}
  }
  \subfigure[Large-scale small objects \label{subfig:scale_small}]{
     \includegraphics[width=0.45\textwidth]{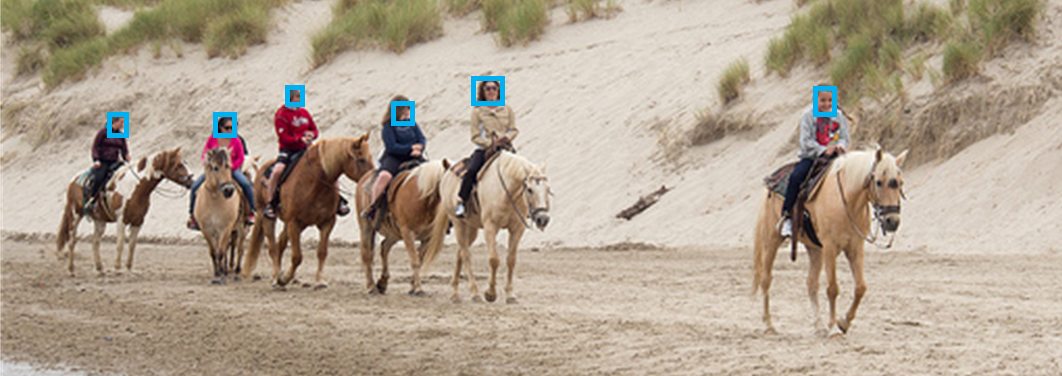}
  }
  \caption{Extreme cases of object detection in images.\label{fig:extreme_cases}}
\end{figure}

Object detection in images is one of the most widely explored tasks in computer vision~\cite{he2017mask,he2016deep}. Existing deep learning approaches to solve this task (e.g., R-CNN~\cite{girshick2014rich} and its variants~\cite{girshick2015fast,ren2015fasterrcnn,he2017mask}) mainly rely on region proposal mechanisms (e.g., region proposal networks (RPNs)) to generate potential bounding boxes in an image and then classify these bounding boxes to achieve object detection. Although such mechanisms can generally achieve a good detection performance under normal circumstances, their recall in a scene with extreme cases (e.g., complex occlusion (Fig.~\ref{subfig:occlusion}), poor illumination  (Fig.~\ref{subfig:lightning}), and large-scale small objects (Fig.~\ref{subfig:scale_small})) is unacceptably low.

Specifically, detecting objects under extreme cases via region proposal mechanisms encounters two challenges: First, the performance of region proposal mechanisms highly depends on the purity of bounding boxes~\cite{guo2016deep}; however, the annotated bounding boxes in extreme cases usually contain much more environment noise than those in normal cases. This inevitably increases the difficulty of model learning and decreases the resulting confidence scores of bounding boxes, which consequently weakens the detection performance. Second, non-maximum suppression (NMS) operations are used in region proposal mechanisms  to select target boxes by setting an intersection over union (IoU) threshold to filter other bounding boxes. However, it is very hard (and sometimes even impossible) to find an appropriate threshold to adapt to the very complex situations in extreme cases.

Motivated by this, in this work, we propose a \emph{weakly supervised multimodal annotation segmentation (WSMA-Seg)} approach, which uses segmentation models to achieve an accurate and robust object detection without NMS. It consists of two phases, namely, a training and a testing phase. In the training phase, WSMA-Seg first converts  weakly supervised bounding box annotations in detection tasks to multi-channel segmentation-like masks, called \emph{multimodal annotations}; then, a segmentation model is trained using multimodal annotations as labels to learn multimodal heatmaps for the training images. In the testing phase, the resulting heatmaps of a given test image are converted into an instance-aware segmentation map based on a pixel-level logic operation; then, a contour tracing operation is conducted to generate contours for objects using the segmentation map; finally, bounding boxes of objects are created as circumscribed quadrilaterals of their corresponding contours.

WSMA-Seg has the following advantages: (i) as an NMS-free solution, WSMA-Seg avoids all hyperparameters related to anchor boxes and NMS; so, the above-mentioned threshold selection problem is also avoided; (ii) the complex occlusion problem can be alleviated by utilizing the topological structure of segmentation-like multimodal annotations; and (iii) multimodal annotations are pixel-level annotations; so, they can describe the objects more accurately and overcome the above-mentioned environment noise problem. 

Furthermore, it is obvious that the performance of the proposed WSMA-Seg approach greatly depends on the segmentation performance of the underlying segmentation model. Therefore, in this work, we further propose a \emph{multi-scale pooling segmentation (MSP-Seg)} model, which is used as the underlying segmentation model of WSMA-Seg to achieve a more accurate segmentation (especially for extreme cases, e.g., very small objects), and consequently enhances the detection accuracy of WSMA-Seg. 



\eat{
The contributions of this paper are briefly  as follows: 
(i) We propose a weakly supervised multimodal annotation segmentation (WSMA-Seg) approach to achieve an accurate and robust object detection without NMS, which is the first anchor-free and NMS-free object detection approach. 
(ii) We propose multimodal annotations to achieve an instance-aware segmentation using weakly supervised bounding boxes; we also 
develop a run-data-based following 
algorithm  to trace contours of objects. 
(iii) We propose a multi-scale pooling segmentation (MSP-Seg) model  to achieve a more accurate segmentation and to enhance the detection accuracy of WSMA-Seg.
(iv) We have conducted extensive experimental studies  on the Rebar Head, WIDER Face, and MS COCO datasets; the results show that the proposed WSMA-Seg approach outperforms the state-of-the-art detectors on all testing datasets.
}

The contributions of this paper are briefly  as follows: 
\begin{itemize}
  \item We propose a weakly supervised multimodal annotation segmentation (WSMA-Seg) approach to achieve an accurate and robust object detection without NMS, which is the first anchor-free and NMS-free object detection approach. 
  \item We propose multimodal annotations to achieve an instance-aware segmentation using weakly supervised bounding boxes; we also 
develop a run-data-based following 
algorithm  to trace contours of objects. 
  \item We propose a multi-scale pooling segmentation (MSP-Seg) model  to achieve a more accurate segmentation and to enhance the detection accuracy of WSMA-Seg.
  \item  We have conducted extensive experimental studies  on the Rebar Head, WIDER Face, and MS COCO datasets; the results show that the proposed WSMA-Seg approach outperforms the state-of-the-art detectors on all testing datasets.
\end{itemize}

\section{Weakly Supervised Multimodal Annotation Segmentation}

\begin{figure}[!t]
  \centering
     \includegraphics[width=0.70\textwidth]{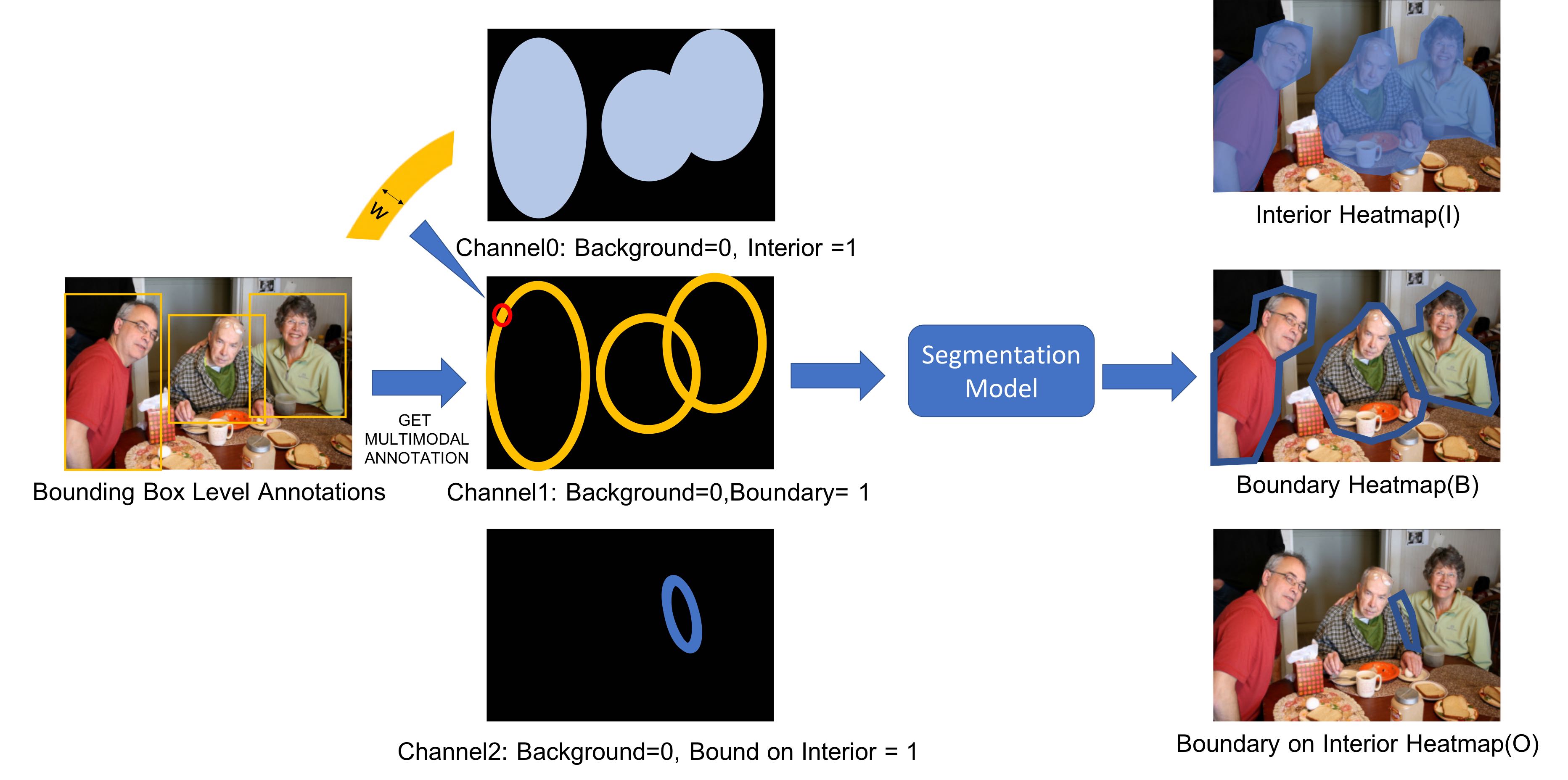}
     \caption{Training phase of WSMA-Seg.\label{fig:training_phrase}}
\end{figure}

\begin{figure}[!t]
\vspace{-0.5em}
  \centering
     \includegraphics[width=0.99\textwidth]{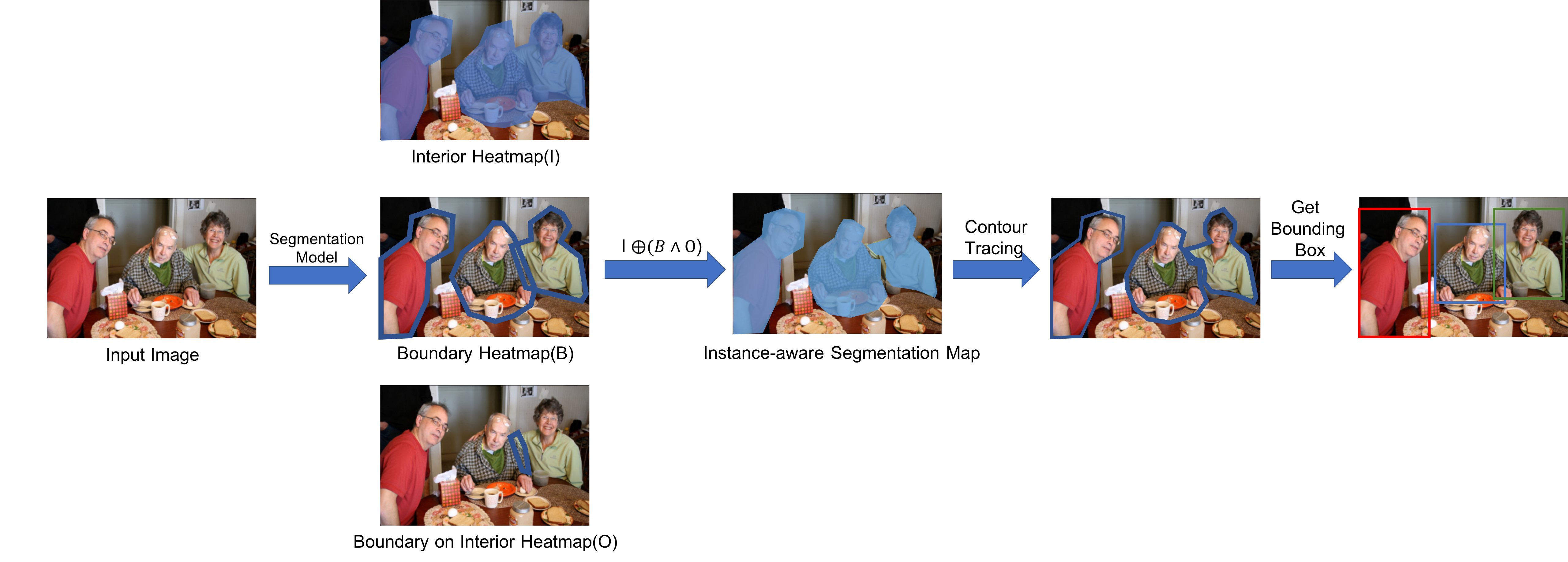}
     \caption{Testing phase of WSMA-Seg.\label{fig:testing_phrase}}
		\vspace{-1.5em}
\end{figure}

In this section, we introduce our approach to object detection using \emph{weakly supervised multimodal annotation segmentation (WSMA-Seg)}. WSMA-Seg generally consists of two phases: a training phase and a testing phase. In the training phase, as shown in Figure~\ref{fig:training_phrase}, WSMA-Seg first converts the weakly supervised bounding box annotations to pixel-level segmentation-like masks with three channels, representing interior, boundary, and boundary on interior masking information, respectively; the resulting annotations are called \emph{multimodal annotations}; then, multimodal annotations are used as labels to train an underlying segmentation model to learn corresponding multimodal heatmaps for the training images. In the testing phase, as shown in Figure~\ref{fig:testing_phrase}, we first send the given testing image into the well-trained segmentation model to obtain multimodal heatmaps; then, the resulting three heatmaps are converted into an instance-aware segmentation map based on a pixel-level logic operation; finally, a contour tracing operation is conducted to generate contours for objects using the segmentation map, and the bounding boxes of objects are created as circumscribed quadrilaterals of their contours. The rest of this section will introduce the main ingredients of WSMA-Seg.

\subsection{Generating Multimodal Annotations}

Pixel-level segmentation annotations are much more representative than bounding box annotations, so they can resolve some extreme cases that are challenging for bounding box annotations. However, creating well-designed pixel-level segmentation masks is very time-consuming, which is about $15$ times of creating bounding box annotations~\cite{lin2014microsoft}. Therefore, in this work, we propose a methodology to automatically convert  bounding box annotations to segmentation-like multimodal annotations, which are pixel-level geometric segmentation-like multichannel annotations. Here, ``geometric segmentation-like'' means that the multimodal annotations are not strict segmentation annotations; rather, they are annotations generated from simple geometries, e.g., inscribed ellipses of bounding boxes. This is motivated by the finding in~\cite{dai2015boxsup} that  pixel-level segmentation information is not fully utilized by segmentation models; we thus believe that  well-designed pixel-level segmentation annotations may not be essential to achieve a reasonable performance; rather,  pixel-level geometric annotations should be enough. Furthermore, to generate a bounding box for each object in the image, an instance-aware segmentation is required; to achieve this, multimodal annotations are designed to have multiple channels to introduce additional information.

Specifically, as shown in Figure~\ref{fig:training_phrase}, multimodal annotations use three channels to represent pixel-level masking information regarding the interior, the boundary, and the boundary on the interior of geometries. These three different pixel-level masks are generated as follows: Given an image with bounding box annotations, we first obtain an inscribed ellipse for each bounding box, then the interior mask (channel $0$) is obtained by setting the values of pixels on the edge of or inside the ellipses to $1$, and setting the values of other pixels to $0$. Then, the boundary mask (channel $1$) is obtained by setting the values of pixels on the edge of or within the inner width $w$ of the ellipses to $1$, and setting the rest to $0$. Similarly, the boundary on the interior mask (channel $2$) is generated by setting the values of pixels on the edge of or within the inner width $w$ of the area of the elliptical overlap to $1$.

\subsection{Multi-Scale Pooling Segmentation}
It is obvious that the performance of the proposed WSMA-Seg approach  greatly depends on the segmentation performance of the underlying segmentation model. Therefore, in this work, we further propose a \emph{multi-scale pooling segmentation (MSP-Seg)} model, which is used as the underlying segmentation model of WSMA-Seg to achieve a more accurate segmentation (especially for extreme cases, e.g., very small objects), and to consequently enhance the detection accuracy of WSMA-Seg. 

\begin{figure*}
  \centering
  \includegraphics[width=0.95\textwidth]{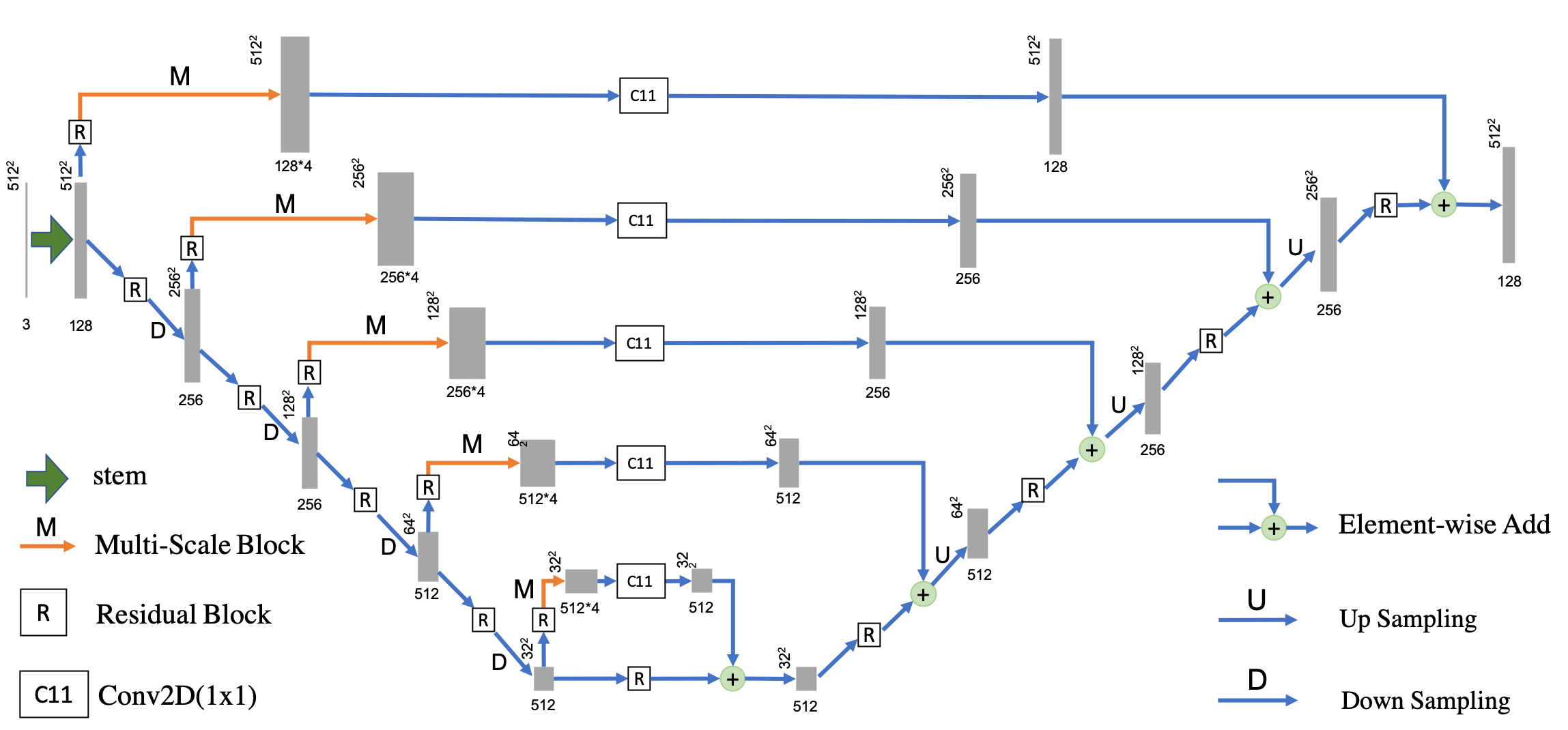}
  \caption{Multi-scale pooling segmentation model.\label{fig:multiscale_model}}
\end{figure*}

\begin{figure}
  \centering
  \includegraphics[width=0.6\textwidth]{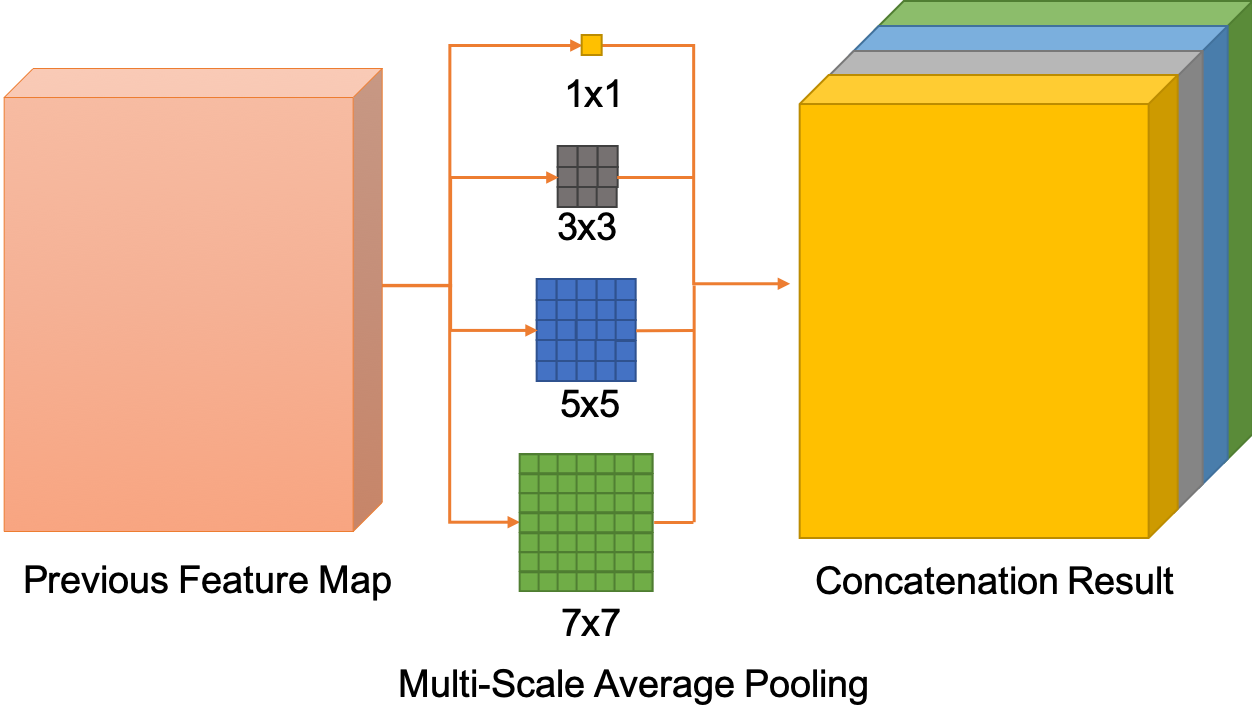}
  \caption{Multi-scale block.\label{fig:fox_block}}
\end{figure}

As shown in Figure~\ref{fig:multiscale_model}, MSP-Seg is an improved segmentation model of Hourglass~\cite{newell2016stacked}. The main improvement of MSP-Seg is to introduce a \emph{multi-scale block} on the skip connections, performing \emph{multi-scale pooling} operations to the output feature maps of residual blocks. Specifically, as shown in Figure~\ref{fig:fox_block}, multi-scale pooling utilizes four pooling kernals with sizes $1\times 1$, $3\times 3$, $5 \times 5$, and $7\times 7$ to simultaneously conduct average pooling operations on the previous feature maps generated by residual blocks on skip connections. Then, four feature maps generated by different pooling channels are concatenated to form a new feature map whose number of channels is four times of the previous feature maps. Here, to ensure that the four feature maps have the same size, the stride is set to $1$, and zero-padding is conducted. Finally, we apply $1\times 1$ convolution  to restore the number of channels, and element-wise addition to merge the feature maps. As shown in Figure~\ref{fig:multiscale_model}, by using multimodal annotations as labels, MSP-Seg is trained to learn three heatmaps for each image, which are called \emph{interior heatmap}, \emph{boundary heatmap}, and \emph{boundary on interior heatmap}, respectively.

Intuitively, multi-scale pooling is capable of enhancing the segmentation accuracy, because it combines features of different scales to obtain more representative feature maps. Please note that, as a highly accurate segmentation model, MSP-Seg can be widely applied to  various segmentation tasks.

\subsection{Object Detection Using Segmentation Results and Contour Tracing}

After obtaining a well-trained segmentation model, we are now able to conduct object detection. As shown in Figure~\ref{fig:testing_phrase}, given a test image as the input of the segmentation model, WSMA-Seg first generates three heatmaps, i.e., interior, boundary, and boundary on interior heatmaps, which are denoted as \emph{I}, \emph{B}, and \emph{O}, respectively. These three heatmaps are then converted to binary heatmaps, where the values of pixels in interested area are set to $1$, and the rest is set to $0$. This conversion is conducted following the approach  in~\cite{suzuki1985topological}. Furthermore, a pixel-level operation, $I \oplus (B \wedge O)$, is used to merge three heatmaps into an instance-aware segmentation map. 

Finally, a contour tracing operation is conducted to generate contours for objects using the instance-aware segmentation map, and the bounding boxes of objects are created as circumscribed quadrilaterals of their contours. One conventional way to trace a contour is to use scan-based-following algorithm~\cite{suzuki1985topological}. However, in the case of a large image with many objects (which is common in detection tasks), scan-based-following algorithm is very time consuming.

\begin{figure*}[!t]
  \centering
     \includegraphics[width=0.85\textwidth]{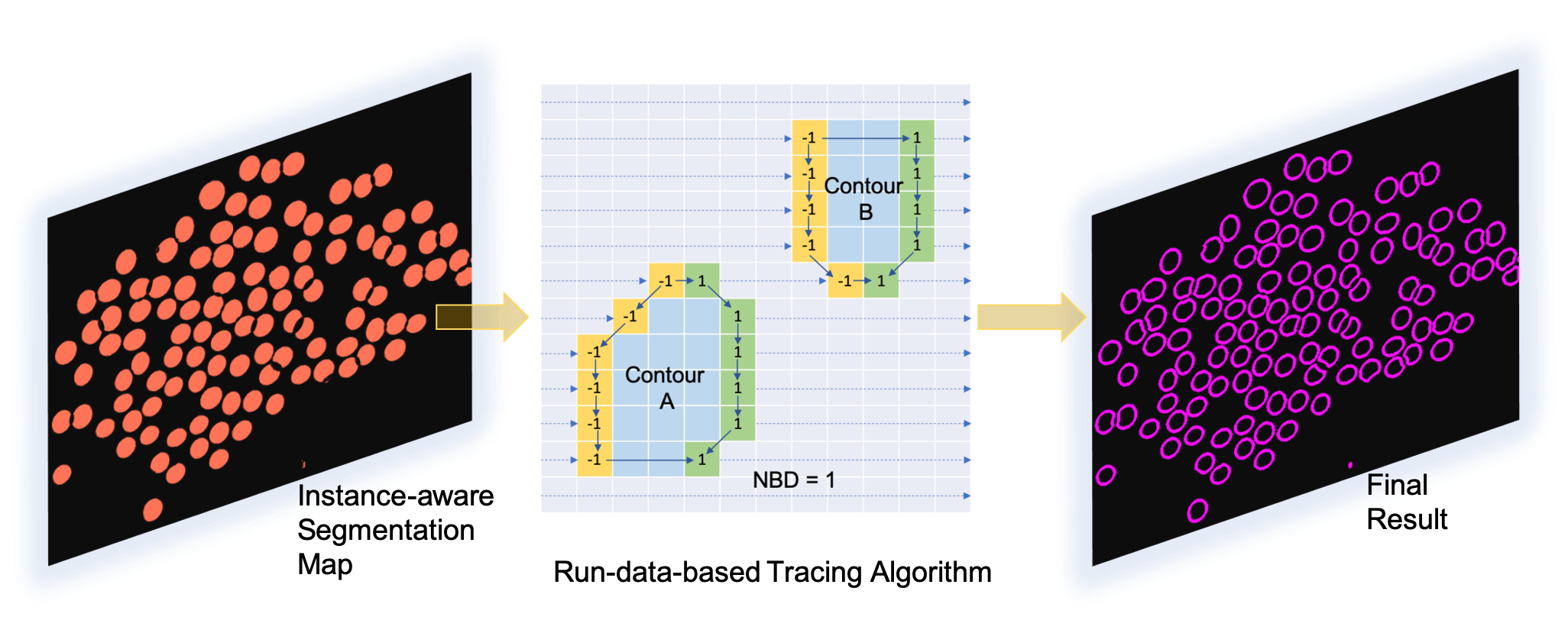}\label{fig:contour_tracing}
  \caption{Contour tracing.}
  \end{figure*}

Therefore, motivated by the work in~\cite{agrawala1977sequential}, we propose a modified run-data-based (RDB) following algorithm, which greatly reduces the time and memory costs of the contour tracing operation. 
Pseudocode of the RDB following algorithm is shown in Algorithm~\ref{rdb} and an example is shown in Figure~\ref{fig:contour_tracing}. Differently from the pixel-following algorithm that requires to scan the entire image to find the starting point and tracing contour pixels along the clockwise direction to generate the results recurrently, the RDB following algorithm only needs to save two lines of pixel values and to scan the whole image once, which significantly reduces the memory consumption and increases the speed.

Specifically, RDB following algorithm first initialize two variables $l_{edge}$ and $r_{edge}$ with \emph{null} value, then scans the binary instance-aware segmentation map row by row from the top-left corner to the bottom-right corner to find contours (lines $1$-$3$). If a pixel's value is $1$ and its left pixel's value is $0$, then this pixel is on the left side of a contour, so it is assigned to $l_{edge}$; similarly, if a pixel's value is $1$ and its right pixel's value is $0$, then this pixel is on the right side of a contour, so it is assigned to $r_{edge}$ (lines $4$-$9$). When both $l_{edge}$ and $r_{edge}$ are found, we check if there exists a pair of $l'_{edge}$ and $r'_{edge}$ on above line whose x-coordinates are the same as or greater/smaller by $1$ than the corresponding x-coordinates of $l_{edge}$ and $r_{edge}$; if so, we add $l_{edge}$ and $r_{edge}$ to the same contour set as $l'_{edge}$ and $r'_{edge}$; otherwise, we create a new contour set and add $l_{edge}$ and $r_{edge}$ to it (lines $10$-$19$).


\begin{algorithm}[!t]
  \caption{Run-data-based following algorithm\label{rdb}}
  \begin{algorithmic}[1]
     \Require
     A binary image with $h$ (height) and $s$ (width).
     \Ensure
     A list of contour sets.
		\State $l_{edge} = null$, $r_{edge} = null$
		\For{$j$ in $[0:h)$}
			\For{$i$ in $[0:s)$}
				\If{ $pixel(i,j)==1$ and $pixel(i-1,j)==0$}
					\State $l_{edge} = Pixel(i,j)$
				\EndIf
				\If{$pixel(i,j)==1$ and $pixel(i+1,j)==0$}
					\State $r_{edge} = Pixel(i,j)$
				\EndIf
				\If{ $r_{edge}$ != $null$ and $l_{edge}$ != $null$}
					\If{there exists a pair of $r’_{edge}$ and $l’_{edge}$ in row $j-1$ and \\ 
					\hspace{45pt} $\left|x_{coordOf}(r’_{edge})- x_{coordOf}(r_{edge})\right| \leq 1$ and\\
					\hspace{45pt} $\left|x_{coordOf}(l’_{edge})- x_{coordOf}(l_{edge})\right| \leq 1$}
						\State Add $l_{edge}$ and $r_{edge}$ to the same contour set as $r’_{edge}$ and $l’_{edge}$
					\Else
						\State Creat a new contour set and add $l_{edge}$ and $r_{edge}$  to it
					\EndIf
					\State $r_{edge}= null$ and $l_{edge} = null$
				\EndIf
			\EndFor
		 \EndFor
  \end{algorithmic}
\end{algorithm}

\section{Experiments}

To show the strength of our proposed WSMA-Seg approach in object detection, extensive experimental studies have been conducted on three benchmark datasets, namely, the Rebar Head\footnote{https://www.datafountain.cn/competitions/332/details}, WIDER Face\footnote{http://mmlab.ie.cuhk.edu.hk/projects/WIDERFace/}, and MS COCO datasets\footnote{http://cocodataset.org/}, each of which containing many extreme cases. The important parameters of WSMA-Seg are as follows: \textit{Stack} is the number of the stacked hourglass networks (see~\cite{newell2016stacked} for more details about hourglass), \textit{Base} is a pre-defined basic number, and the number of channels is always an integer  multiple of Base, and \textit{Depth} is the number of down-samplings. \textit{Stem} represents three consecutive $3\times 3$ convolution operations with stride $=1$ before the first stack.

\subsection{Rebar Head Detection}

\begin{figure}[!t]
  \centering
  \includegraphics[width=0.45\textwidth]{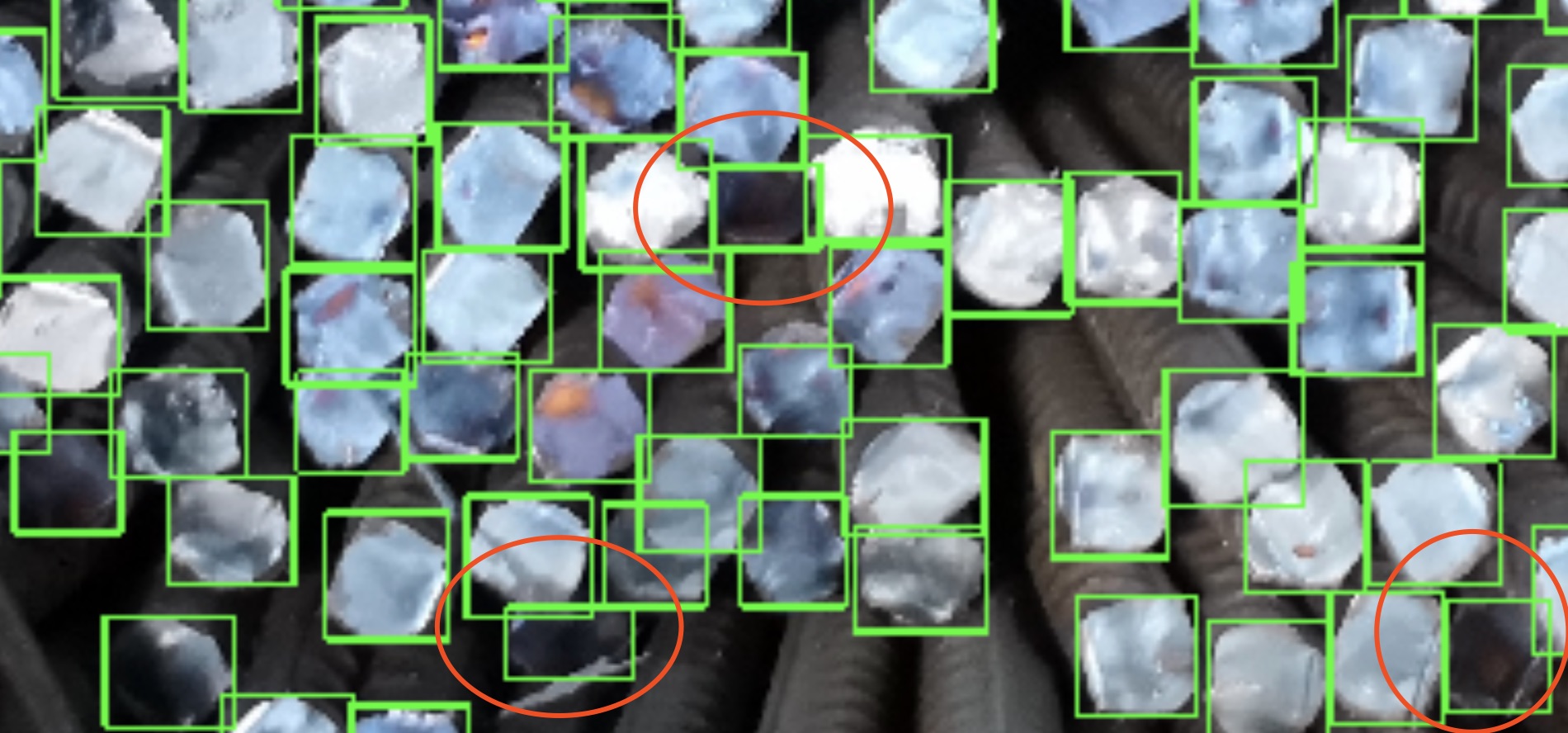}
  \caption{An example of complex occlusion in the Rebar Head dataset.\label{fig:rebar_example}}
\end{figure}

We first conduct experiments on the Rebar Head detection dataset, which consists of $250$ training images (including a total of $30942$ rebar heads) and $200$ testing images. The orignal resolution of the whole image is $2000 \times 2666$. Performing object detection on this dataset is very challenging, because it only contains a few training samples  and also encounters very severe occlusion situations (see Figure~\ref{fig:rebar_example}). 
In addition, the target rebar heads are very small: the average area of each box is $7,000$ pixels, taking up only 0.13\% of the whole image. The images are also poorly annotated and rich in diverse illuminations. 

Two state-of-the-art anchor-based models, Faster R-CNN~\cite{ren2015fasterrcnn} and Cascade R-CNN~\cite{cai2018cascade}, are selected as the baselines. Table~\ref{table:rebar} shows the detection performances of our proposed WSMA-Seg and baselines on this dataset.
As shown in Table~\ref{table:rebar}, our proposed method with $Stack=2$, $Base=40$, $Depth=5$ has achieved the best performance among all solutions in terms of F1 Score. In addition, the number of parameters needed for WSMA-Seg is much less than the baselines (only $1/7$ of Cascade RCNN and $1/4$ of Faster RCNN), while the number of training epochs for WSMA-Seg is also less than those of the baselines. Therefore, we can conclude that, compared to the state-of-the-art baselines, WSMA-Seg is much simpler, more effective, and more efficient.

\begin{table*}[!t]
  \caption{Detection performances of WSMA-Seg and baselines on the Rebar Head dataset.\label{table:rebar}}
  
  \smallskip 
  \centering
  \begin{tabular}{|c|c|c|c|c|c|c|}
  \hline
  Method&\#parms&Epoch&F1 Score\\ \hline
  Faster RCNN&23.2M&100&98.30\%\\\hline
  Cascade RCNN~\cite{cai2018cascade}&42.1M&100&98.70\%\\\hline
  WSMA-Seg(stack=1,base=72,depth=3)&6.1M&70&94.27\%\\\hline
  WSMA-Seg(stack=2,base=40,depth=5)&5.8M&70&\textbf{98.83\%}\\\hline
  WSMA-Seg(stack=4,base=28,depth=5)&5.7M&70&96.26\%\\\hline
  \end{tabular}
\end{table*}

\subsection{WIDER Face Detection}
We further conduct experiments on the WIDER Face detection dataset~\cite{yang2016wider}, which consists of $32,203$ images and $393,703$ faces. Face detections in this dataset are extremely challenging due to a high degree of variability in scale, pose, and occlusion. WIDER Face results in a much lower detection accuracy compared to other face detection datasets. WIDER Face has defined three levels of difficulties (i.e., \textit{Easy}, \textit{Medium}, and \textit{Hard}),  based on the detection accuracies of EdgeBox~\cite{zitnick2014edge}. 
Furthermore, the dataset also treats occlusion as an additional attribute and is partitioned into three categories: \textit{no occlusion}, \textit{partial occlusion}, and \textit{heavy occlusion}. Specifically, a face is categorized as \textit{partial occlusion} when 1\% to 30\% of the total face area is occluded, and a face with the occluded area over 30\% is categorized as \textit{heavy occlusion}. The size of the training set is $12879$, that of the validation set is $3226$, and that of the testing set is $16098$.

\begin{figure*}
  \centering
  \includegraphics[width=0.9\textwidth]{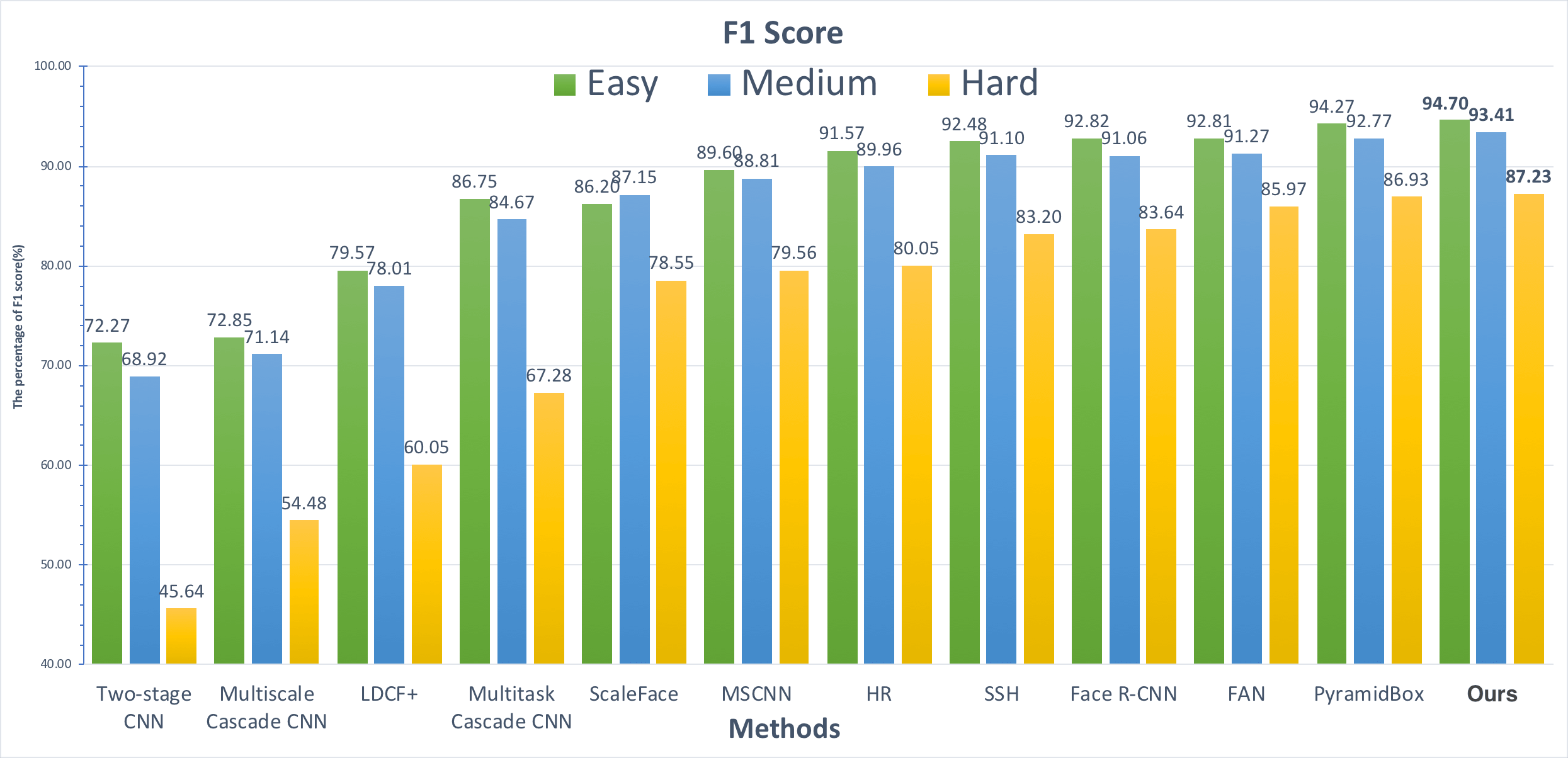}
  \caption{F1 scores on the WIDER Face dataset.\label{table:widerface_f1}}
\end{figure*}

Twelve state-of-the-art approaches are selected as baselines, namely, Two-stage CNN, Cascade R-CNN, and LDCF+\cite{ohn2016boost}, multitask Cascade CNN~\cite{zhang2016joint}, ScaleFace~\cite{yang2017face}, MSCNN~\cite{cai2016unified}, HR~\cite{hu2017finding}, Face R-CNN~\cite{wang2017detecting}, Face Attention Networks~\cite{wang2017face}, and PyramidBox~\cite{tang2018pyramidbox}.
The experimental results in terms of F1 score are shown in Table~\ref{table:widerface_f1}.
The results show that our proposed WSMA-Seg outperforms the state-of-the-art baselines in all three categories, reaching $94.70$, $93.41$, and $87.23$ in Easy, Medium, and Hard categories, respectively.


\eat{
\begin{table*}[h]
  \centering
  \caption{Ablation study on Widerface Val Dataset\label{table:widerface_ablation}}
  \begin{small}
    \begin{tabular}{|c|c|c|c|c|c|c|c|c|c|c|c|}
      \hline
         Stem&Stack&Base&Depth&\#params&Time&Epoch&Easy&Medium&Hard\\ \hline
         \Checkmark&2&40&5&6.08&32.19&65&\textbf{94.70\%}&\textbf{93.41\%}&\textbf{87.23\%}\\\hline
         &2&40&5&5.78&29.47&65&91.45\%&90.50\%&82.86\%\\\hline
         \Checkmark&2&40&5&6.07&32.16&65&93.83\%&92.56\%&84.33\%\\\hline
         \Checkmark&2&40&5&5.55&30.52&65&89.57\%&88.85\%&79.56\%\\\hline
         \Checkmark&4&28&5&5.68&34.58&65&90.53\%&90.01\%&82.54\%\\\hline
         \Checkmark&1&72&3&6.13&32.17&65&90.67\%&89.86\%&82.31\%\\\hline
      \end{tabular}
  \end{small}
\end{table*}
}

\subsection{MS COCO Detection}
Finally, we conduct experimental studies on the MS COCO detection dataset~\cite{lin2014microsoft}, which is one of the most popular large-scale detection datasets. 
Our results are obtained using the test-dev split (20k images) with a host of the detection method. We have constructed the training set with $82081$ samples, the validation set with $40137$ samples, and the testing set with $20288$ samples. 
We use the metrics as used in \cite{lin2014microsoft} to characterize the performance. Four types of metrics are defined and described as follows:

\begin{small}
\begin{itemize}
  \item \textbf{Average Precision (AP):}
  \begin{itemize}[noitemsep]
     \item $AP$: AP at IoU=.50:.05:.95 (primary challenge metric)
     \item $AP^{.50}$: AP at IoU=.50 (PASCAL VOC metric)
     \item AP at IoU=.75 (strict metric)
  \end{itemize}
  \item \textbf{AP Across Scales:}
  \begin{itemize}[noitemsep]
     \item $AP^{s}$: AP for small objects: area  $<32^2$
     \item $AP^{m}$: AP for medium objects: $32^2<$  area  $<96^2$
     \item $AP^{l}$ : AP for large objects: area  $>96^2$
  \end{itemize}
  \item \textbf{Average Recall (AR):}
  \begin{itemize}[noitemsep]
     \item $AR^{1}$ : AR given 1 detection per image
     \item $AR^{10}$ : AR given 10 detections per image
     \item $AR^{100}$ : AR given 100 detections per image
  \end{itemize}
  \item \textbf{AR Across Scales:}
  \begin{itemize}[noitemsep]
     \item $AR^{s}$ : AR for small objects: area  $<32^2$
     \item $AR^{m}$ : AR for medium objects: $32^2<$ area $< 96^2$
     \item $AR^{l}$ : AR for large objects: area$>96^2$
  \end{itemize}
\end{itemize}
\end{small}
Seven state-of-the-art solutions are selected as baselines, and the experimental results for four types of metrics are shown in Tables~\ref{coco_precision} and~\ref{coco_recall}. The results show that our WSMA-Seg approach outperforms all state-of-the-art baselines in terms of most metrics, including the most challenging metrics, $AP$, $AP^s$, $AR^1$, and $AR^s$. For the other metrics, the performance of our proposed approach is also close to those of the best baselines. This proves that the proposed WSMA-Seg
approach generally achieves more accurate and robust object detection than the state-of-the-art approaches without NMS.

\begin{table*}[!ht]
   \centering
   \vspace{-0.5em}
	\caption{Average precisions of WSAM-Seg and baselines on MS COCO (test-dev)\label{coco_precision}}
	\begin{small}
   \begin{tabular}{|l|l|c|c|c|c|c|c|}
   \hline
   Method&Backbone& $AP$ & $AP^{50}$ &$AP^{75}$&$AP^{s}$& $AP^{m}$& $AP^{l}$ \\\hline
   DSOD300&DS/64-192-48-1&29.3&47.3&30.6&9.4&31.5&47\\\hline
   SSD513&ResNet-101&31.2&50.4&33.3&10.2&34.5&49.8\\\hline
   DSSD513&ResNet-101&33.2&53.3&35.2&13.0&35.4&51.1\\\hline
   DeNet&ResNet-101&33.8&53.4&36.1&12.3&36.1&50.8\\\hline
   CoupleNet&ResNet-101&34.4&54.8&37.2&13.4&38.1&50.8\\\hline
   Faster R-CNN w/ TDM&Inception-ResNet-v2&36.8&57.7&39.2&16.2&39.8&\textbf{52.1}\\\hline
   CornerNet511&Hourglass-52&37.8&53.7&40.1&17.0&39.0&50.5\\\hline
   WSMA-Seg&MSP-Seg&\textbf{38.1}&\textbf{58.2}&\textbf{40.7}&\textbf{22.5}&\textbf{41.0}&51.9\\\hline
  \end{tabular}
	\end{small}
\end{table*}

\begin{table*}[!ht]
   \centering
   \vspace{-0.5em}
	\caption{Average recalls of WSAM-Seg and baselines on MS COCO (test-dev)\label{coco_recall}}
	\begin{small}
   \begin{tabular}{|l|l|c|c|c|c|c|c|}
   \hline
   Method&Backbone& $AR^{1}$& $AR^{10}$& $AR^{100}$&$AR^{s}$& $AR^{m}$& $AR^{l}$ \\\hline
   DSOD300&DS/64-192-48-1&27.3&40.7&43&16.7&47.1&65\\\hline
   SSD513&ResNet-101&28.3&42.1&44.4&17.6&49.2&65.8\\\hline
   DSSD513&ResNet-101&28.9&43.5&46.2&21.8&49.1&66.4\\\hline
   DeNet&ResNet-101&29.6&42.6&43.5&19.2&46.9&64.3\\\hline
   CoupleNet&ResNet-101&30.0&45.0&46.4&20.7&53.1&68.5\\\hline
   Faster R-CNN w/ TDM&Inception-ResNet-v2&31.6&49.3&51.9&28.1&56.6&71.1\\\hline
   CornerNet511&Hourglass-52&33.9&\textbf{52.3}&57.0&35.0&\textbf{59.3}&\textbf{74.7}\\\hline
   WSMA-Seg&MSP-Seg&\textbf{35.2}&52.1&\textbf{57.8}&\textbf{36.1}&58.4&73.2\\\hline
  \end{tabular}
	\end{small}
\end{table*}
	
\section{Conclusion}
In this work, we have proposed a novel approach to object detection in images, called weakly su\-per\-vised multimodal annotation segmentation (WSMA-Seg), which is anchor-free and NMS-free. We observed that  NMS is one of the bottlenecks of existing deep learning 
approaches to object detection in images. The need to tune hyperparameters on NMS has seriously hindered the scalability of high-performance detection frameworks. Therefore, to realize WSMA-Seg, we proposed to use multimodal annotations to achieve an instance-aware segmentation based on weakly supervised bounding boxes, and developed a run-data-based following algorithm to trace contours of objects. In addition, a multi-scale pooling segmentation (MSP-Seg) model was proposed as the underlying segmentation model of WSMA-Seg to achieve a more accurate segmentation and to enhance the detection accuracy of WSMA-Seg. Experimental results on multiple datasets concluded that the proposed WSMA-Seg approach is superior to the state-of-the-art detectors.
{\small
\bibliographystyle{IEEEtran}
\bibliography{egbib}
}
\end{document}